\documentclass[lettersize,journal]{IEEEtran}

\usepackage{amsmath, amssymb, amsfonts, mathtools, mathrsfs, amsthm,bm}
\usepackage{newtxtext,newtxmath}

\usepackage[ruled,vlined,linesnumbered]{algorithm2e}

\usepackage{booktabs}
\usepackage{multirow}
\usepackage{tabularx}
\usepackage{array}
\usepackage{colortbl}
\usepackage{xcolor}
\usepackage[table]{xcolor}
\usepackage{graphicx}
\usepackage{float}
\usepackage[caption=false,font=normalsize,labelfont=sf,textfont=sf]{subfig}

\usepackage{hyperref}
\usepackage{orcidlink}
\usepackage{textcomp}
\usepackage{url}
\usepackage{verbatim}
\usepackage{booktabs}

\usepackage{booktabs}
\usepackage{tabularx}
\usepackage{multirow}
\usepackage{newtxtext,newtxmath} 

\usepackage{tikz}
\usetikzlibrary{shapes.geometric, arrows}

\tikzstyle{startstop} = [rectangle, rounded corners, minimum width=3cm, minimum height=1cm,text centered, draw=black, fill=red!20]
\tikzstyle{process} = [rectangle, minimum width=3.5cm, minimum height=1cm, text centered, draw=black, fill=blue!20]
\tikzstyle{decision} = [diamond, minimum width=3cm, minimum height=1cm, text centered, draw=black, fill=green!20]
\tikzstyle{arrow} = [thick,->,>=stealth]

\usepackage{enumitem}

\usepackage{stfloats}
\usepackage{balance}
\def\BibTeX{{\rm B\kern-.05em{\sc i\kern-.025em b}\kern-.08em
		T\kern-.1667em\lower.7ex\hbox{E}\kern-.125emX}}
\begin{document}

\title{PiDR: Physics-Informed Inertial Dead Reckoning for Autonomous Platforms}

\author{
	Arup Kumar Sahoo \orcidlink{0000-0003-4515-7434} 
	\thanks{}
	and Itzik Klein \orcidlink{0000-0001-7846-0654} 
	\thanks{ The authors are with the Autonomous Navigation and Sensor Fusion Lab (ANSFL), the Hatter Department of Marine Technologies,
		Charney School of Marine Sciences, University of Haifa, Haifa 3498838,
		Israel. (e-mail: asahoo@campus.haifa.ac.il, kitzik@univ.haifa.ac.il } }


\maketitle

\begin{abstract}
\noindent
A fundamental requirement for full autonomy is the ability to sustain accurate navigation in the absence of external data, such as GNSS signals or visual information. In these challenging environments, the platform must rely exclusively on inertial sensors, leading to pure inertial navigation. However, the inherent noise and other error terms of the inertial sensors in such real-world scenarios will cause the navigation solution to drift over time. Although conventional deep-learning models have emerged as a possible approach to inertial navigation, they are inherently black-box in nature. Furthermore, they struggle to learn effectively with limited supervised sensor data and often fail to preserve physical principles. To address these limitations, we propose PiDR, a physics-informed inertial dead-reckoning framework for autonomous platforms in situations of pure inertial navigation. PiDR offers transparency by explicitly integrating inertial navigation principles into the network training process through the physics-informed residual component. PiDR plays a crucial role in mitigating abrupt trajectory deviations even under limited or sparse supervision. To further enhance estimation accuracy, an extended Kalman filter is used as a refinement layer, treating the PiDR outputs as external measurements to update the filter.
We evaluated our PiDR on real-world datasets collected by a mobile robot and an autonomous underwater vehicle. We obtained an average of a minimum 55\% positioning improvement in both datasets, demonstrating the ability of the proposed model to generalize different platforms operating in various environments and dynamics. Thus, our model offers a robust yet effective architecture and can be deployed on resource-constrained platforms, enabling real-time pure inertial navigation in adverse scenarios.

\end{abstract}


\begin{table}[htbp]
	\centering
	{\renewcommand{\thetable}{} 
    \captionsetup{labelformat=empty}
		\caption{\textbf{List of Abbreviations}} 
	}
\addtocounter{table}{-1}
	\renewcommand{\arraystretch}{1.2}
	\begin{tabular}{@{}ll@{}} 
		\toprule
		\textbf{Abbreviation} & \textbf{Definition} \\
		\midrule
		
		INS    & Inertial Navigation System \\
	IDR    & Inertial Dead Reckoning \\
	IMU    & Inertial Measurement Unit \\
	GNSS   & Global Navigation Satellite System \\
	RTK    & Real-Time Kinematic \\
	NED    & North-East-Down  \\
	ECEF   & Earth-Centered Earth-Fixed \\
	PINN   & Physics-informed Neural Network \\
    EKF & Extended Kalman Filter \\
	GT     & Ground Truth \\
	AUV & Autonomous Underwater Vehicle \\
	ATE    & Absolute Trajectory Error \\
	MATE   & Mean Absolute Trajectory Error \\
	MSE    & Mean Squared Error \\
	PRMSE  & Position Root Mean Squared Error \\
	TDE    & Total Distance Error \\
	FDE    & Final Distance Error\\
    XAI    & Explainable Artificial Intelligence\\
		\bottomrule
	\end{tabular}
\end{table}

\section{Introduction}

A fundamental requirement for full autonomy in mobile robots is accurate navigation, even in situations where satellite navigation (outdoors) or cameras (indoors) are unavailable. In such adverse scenarios, the platform must rely exclusively on inertial sensors, leading to pure inertial navigation~\cite{titterton2004strapdown, farrell2008aided, groves2013book}. 
The inertial navigation solution (INS) estimates the position, velocity, and orientation of autonomous platforms by measuring their specific force (excluding gravity) and angular velocity vectors in GNSS-denied environments. This is achieved through using an inertial measurement unit (IMU), which consists of tri-axial accelerometers and gyroscopes, typically arranged in orthogonal triads.

\noindent The primary limitation of the INS is solution drift during pure inertial operation. During integration, the inherent instrumental noise and error terms of inertial sensors penetrate the navigation solution. Consequently, irrespective of sensor grade, the INS solution will drift over time. To circumvent drift in situations of pure inertial navigation (e.g., in GNSS-defined environments), it can be fused with information aiding, where information from platform dynamics or environments is translated into a pseudo-measurement~\cite{engelsman2023information, hwang2021identification, ye2025comprehensive}. Another approach is to use multiple IMUs and utilize the constraints between the sensors to mitigate inertial drift~\cite{nilsson2016inertial, libero2024augmented}. Periodic trajectories were suggested as a means to increase the inertial signal-to-noise ratio. This enabled model-based and deep-learning (DL) approaches to regress position displacements for quadrotors~\cite{shurin2020qdr, bergantin2023indoor} and mobile robots~\cite{etzion2023morpi}. 
Machine learning and DL approaches demonstrate improvements over model-based approaches in various inertial tasks as summarized in recent survey papers~\cite{li2021inertial,roy2021survey,golroudbari2023recent,cohen2024inertial,chen2024deep}. RoNIN~\cite{herath2020ronin}, AI-IMU~\cite{9035481}, RNIN-VIO~\cite{RNIN-VIO} and MoRPINet~\cite{etzion2025snake} are typical examples of data-driven models.

\noindent Conventional neural networks' black-box nature limits their explainability in safety-critical applications such as navigation~\cite{bengio2017deep}. Moreover, to facilitate the training and validation of DL algorithms for autonomous platforms, a large quantity of recorded sensor data is needed. In many situations, such datasets are difficult to acquire or not publicly available. Nonetheless, these purely data-driven models suffer from poor generalization across devices, users, and patterns and also violate the known physical constraints of inertial navigation~\cite{xu2022physics}.

\noindent 
In response to the limitations of black-box models, explainable artificial intelligence (XAI)~\cite{chamola2023review} has gained increasing attention for safety-critical applications such as medical diagnosis, defence, finance, and autonomous vehicles. One such XAI technique is physics-informed neural networks (PINNs). Originally proposed by Raissi \textit{et al.} \cite{raissi2019physics}, PINNs embed underlying physical laws, typically expressed as partial differential equations, directly into the objective function. It successfully mitigates the shortcomings of existing black-box models and also retains the representation power of deep neural networks (DNN) \cite{sahoo2025cl,kumar2023physics, sahoo2024unsupervised}. 
The PINN paradigm has recently been extended to inertial navigation and DR problems. By embedding the fundamental laws of strapdown inertial navigation equations of motion as differential constraints alongside ground-truth (GT), PINN thereby improves interpretability, physical consistency, and generalization.

\noindent 
Xu \textit{et al.} \cite{xu2022physics} applied PINNs to the inertial dynamics of unmanned surface vehicles by enforcing surge/sway residuals for drift mitigation. 
It was followed by Chenkai \textit{et al.} \cite{tan2023vehicle}, who proposed a manifold-aware vehicle state estimation.  
In a different domain, SSPINNpose~\cite{gambietzself} was developed to model self-supervised human movement dynamics using physics-informed learning.
Despite these developments, relatively few studies have explored the application of PINNs to inertial navigation.
Recently, Sahoo and Klein~\cite{sahoo2025morpi} proposed MoRPI-PINN, an information-aided framework to reduce the inertial drift. It embeds the 2D-INS equations of motion into the training process of a neural network, typically focusing on periodic trajectories by mobile robots.
Collectively, the above research demonstrate PINNs' potential for GNSS-denied environments by balancing data fidelity with physics-based residuals. However, most existing PINN formulations treat the entire trajectory as a single spatiotemporal domain, leading to high memory requirements and difficulties in real-time implementation on resource-constrained platforms.

\noindent  It is hypothesized that, during the training, data loss ensures that the predicted trajectories are not too far from the real-world dynamics, and the physics loss ensures that the predicted solutions follow the underlying motion dynamics of the universe. This enables the autonomous platforms to maintain reliable navigation performance even under sensor noise, drift, and in the absence of external aiding.

\noindent
To address this gap in situations of pure inertial navigation, we developed PiDR, a physics-informed framework for autonomous inertial navigation in GNSS-denied environments and in situations where other external updates are unavailable. PiDR receives inertial data and provides the platform’s navigation solution, namely the position, velocity, and orientation. 

\noindent The contributions of this research are:

\begin{enumerate}

	\item Formulation of a physics-informed inertial dead-reckoning (PiDR) model that embeds strapdown inertial navigation as physical constraints directly into a DNN architecture for pure inertial navigation in GNSS-denied operations.
	\item A transparent learning framework that addresses the limited interpretability and explainability of existing learning-based navigation methods and mitigates the inertial drift.
    \item {Integration of an EKF-based refinement stage, treating the PiDR outputs as external measurements to update the filter.}
	\item Cross-validation of our PiDR and PiDR-EKF approaches was made on different platforms operating in different environments and dynamics.
\end{enumerate}

\noindent Comprehensive experimental validation on multiple autonomous platforms (mobile robot and autonomous underwater vehicle) using real recorded datasets of 229 [min] was made. We demonstrated that PiDR achieves an average of a minimum 55\% improvement in positioning accuracy over existing model-based and learning-based approaches.

\noindent
The rest of the paper is organized as follows: Section \ref{sec:Prelim} introduces the preliminaries and mathematical background of strapdown inertial navigation. Section~\ref{sec: problem} gives our proposed approach. Section~\ref{sec: analysis} verifies the effectiveness and superiority of
the proposed method in two platforms and provides an in-depth
analysis of the interpretability of the proposed method. Lastly, 
Section~\ref{sec:conclusion} concludes this work.

\vspace{0.2cm}

\section{Inertial Navigation Preliminaries} \label{sec:Prelim}
\noindent  This section begins with the definition of coordinate frames and reference conventions adopted in the navigation formulation. The classical strapdown inertial navigation equations are then derived to establish the baseline INS model. 

\vspace{0.1cm}
\subsection{Reference Frames}
\noindent In strapdown navigation, states must be consistently transformed between multiple reference frames due to sensor placement and dynamics. In this work we use the following reference frames \cite{titterton2004strapdown, farrell2008aided, groves2013book}:

\begin{itemize}
	\item \textbf{b-frame}: The body-fixed reference frame attached to the vehicle.  
	Its axes define the forward ($x$), right ($y$), and down ($z$) directions of the platform.  
	IMU measurements, such as specific force and angular rates, are expressed in this frame, as it is assumed that the inertial sensor sensitive axes align with the b-frame.
	
	\item \textbf{n-frame}: The navigation frame's origin is the physical location where the navigation state is being determined. In the north-east-down (NED) coordinate frame, the x-axis points towards the geodetic north, the z-axis is on the local vertical pointing down, and the y-axis completes a right-handed orthogonal frame.

\end{itemize}
\vspace{0.1cm}
\subsection{Inertial Dead Reckoning}

\noindent
The platform's position $\mathbf{p}^n \in \mathbb{R}^3$ is parameterized by latitude ($\varphi$), longitude ($\lambda$), and height ($h$). 
The kinematic relationship between the position states and the velocity components expressed in the navigation frame is~\cite{groves2013book, jekeli2023inertial}:
\begin{equation}
	\dot{\mathbf{p}}^n =
	\begin{bmatrix}
		\dot{\varphi} \\
		\dot{\lambda} \\
		\dot{h}
	\end{bmatrix}
	=
	\begin{bmatrix}
		\frac{1}{R_M+h} & 0 & 0 \\
		0 & \frac{1}{(R_N+h)\cos\varphi} & 0 \\
		0 & 0 & -1
	\end{bmatrix}
	\begin{bmatrix}
		v_N \\
		v_E \\
		v_D
	\end{bmatrix}
	\equiv \mathbf{D}(\varphi, h)\,\mathbf{v}^n,
	\label{eq:pos}
\end{equation}
where $\mathbf{v}^n = [v_N, v_E, v_D]^T$ is the velocity vector expressed in the n-frame. Here, $R_M$ and $R_N$ denote the meridian and transverse radii of curvature of the reference ellipsoid and are defined as
\begin{equation}
	R_M = \frac{a(1-e^2)}{(1-e^2\sin^2\varphi)^{3/2}}, \qquad
	R_N = \frac{a}{\sqrt{1-e^2\sin^2\varphi}},
	\label{eq:curvature_radii}
\end{equation}
where $a$ is the Earth's semi-major axis and $e$ its eccentricity.

\noindent
The velocity rate of change expressed in the n-frame is:
\begin{equation}
	\dot{\mathbf{v}}^n 
	= \mathbf{C}_b^n \mathbf{f}_{ib}^b
	- \big(2\boldsymbol{\omega}_{ie}^n + \boldsymbol{\omega}_{en}^n \big) \times \mathbf{v}^n
	+ \mathbf{g}^n,
	\label{eq:vel}
\end{equation}
where $\mathbf{f}_{ib}^b$ is the specific force vector expressed in the b-frame,  
$\boldsymbol{\omega}_{ie}^n$ is the Earth rotation rate expressed in the n-frame, 
$\boldsymbol{\omega}_{en}^n$ is the transport rate due to vehicle motion over the Earth's curved surface, and
$\mathbf{g}^n$ is the gravity vector expressed in the n-frame. 
These vectors are defined as:

\begin{align}
	\mathbf{f}_{ib}^b &= 
	\begin{bmatrix} f_x & f_y & f_z \end{bmatrix}^\top, \\[4pt]
	\boldsymbol{\omega}_{ie}^n &= 
	\begin{bmatrix} \omega_e \cos\varphi & 0 & -\omega_e \sin\varphi \end{bmatrix}^\top,
	\label{eq:earth_rate_n} \\[4pt]
	\boldsymbol{\omega}_{en}^n &= 
	\begin{bmatrix}
		\frac{v_E}{N+h} & -\frac{v_N}{M+h} & -\frac{v_E \tan\varphi}{N+h}
	\end{bmatrix}^\top,
	\label{eq:transport_rate} \\[4pt]
	\mathbf{g}^n &= 
	\begin{bmatrix} 0 & 0 & g(\varphi,h) \end{bmatrix}^\top,
\end{align}
where $f_x, f_y, f_z$ are the components of the specific force vector measured by accelerometer, and 
$\omega_e=7.2921158$ $rad/s$ is the magnitude of the rotation rate of the Earth.

\noindent
The direction cosine matrix (DCM) $\mathbf{C}_b^n \in SO(3)$, which transforms vectors from the b-frame to the local n-frame, evolves over time as:
\begin{equation}
	\dot{\mathbf{C}}_b^n
	=
	\mathbf{C}_b^n \boldsymbol{\Omega}_{ib}^b
	-
	\big( \boldsymbol{\Omega}_{ie}^n + \boldsymbol{\Omega}_{en}^n \big)
	\mathbf{C}_b^n,
	\label{eq:ori}
\end{equation}
where $\boldsymbol{\Omega}_{ib}^b$ is the skew-symmetric matrix of the angular rate $\boldsymbol{\omega}_{ib}^b = [\omega_x, \omega_y, \omega_z]^\top$ measured by the gyroscopes in the b-frame. It is defined as
\begin{equation}
	\boldsymbol{\Omega}_{ib}^b (\boldsymbol{\omega}) \triangleq
	\begin{bmatrix}
		0       & -\omega_z &  \omega_y \\
		\omega_z & 0        & -\omega_x \\
		-\omega_y & \omega_x & 0
	\end{bmatrix}.
\end{equation}
When parameterized by the Euler angles $\boldsymbol{\eta} = [\phi, \theta, \psi]^\top$, the DCM is given by \cite{shin2002accuracy}

\begin{equation}
	\mathbf{C}_b^n =
	\begin{bmatrix}
		c\theta c\psi & c\theta s\psi & -s\theta \\
		s\phi s\theta c\psi - c\phi s\psi &
		s\phi s\theta s\psi + c\phi c\psi &
		s\phi c\theta \\
		c\phi s\theta c\psi + s\phi s\psi &
		c\phi s\theta s\psi - s\phi c\psi &
		c\phi c\theta
	\end{bmatrix},
	\label{eq:DCM_Euler}
\end{equation}
where $c(\cdot)=\cos(\cdot)$, and $s(\cdot)=\sin(\cdot)$.

\vspace{0.2cm}
\section{Proposed Approach} \label{sec: problem}

\begin{figure*}[!t]
    \centering
    \includegraphics[width=0.95\textwidth]{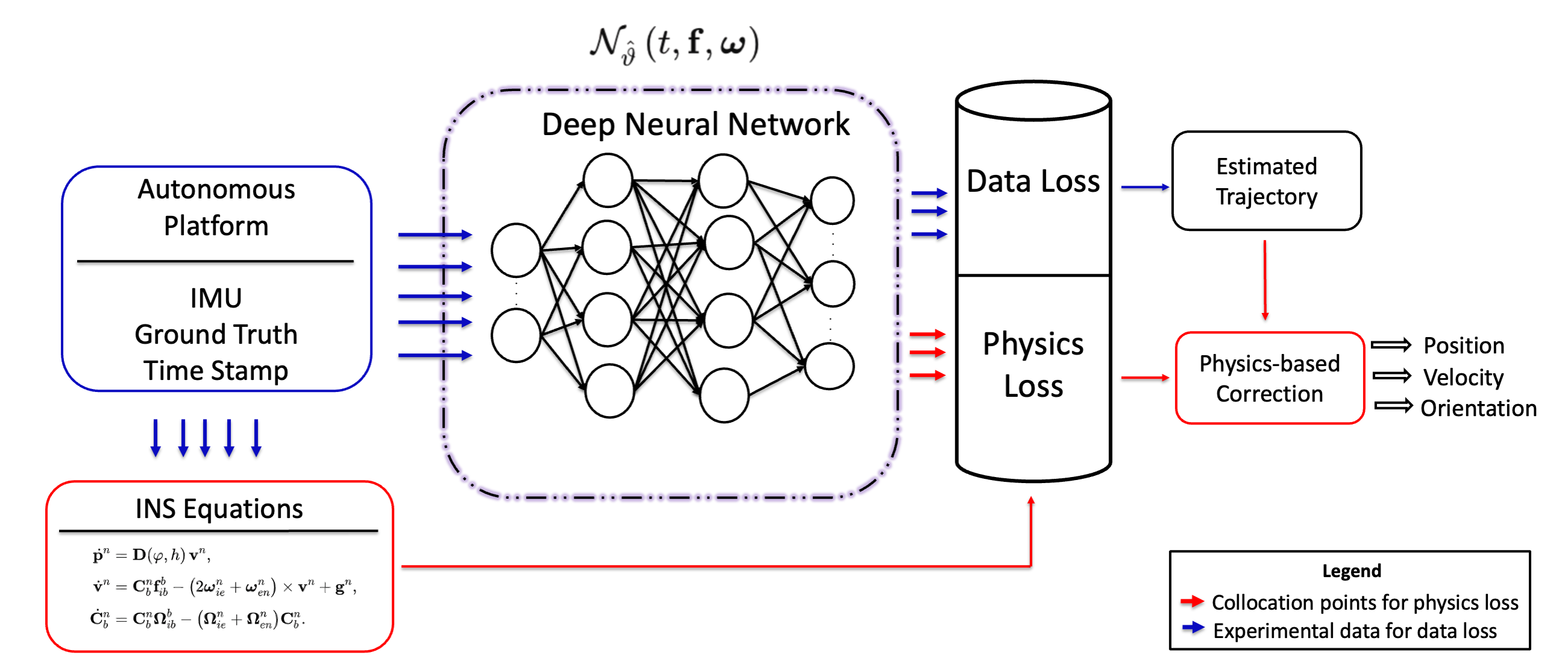}
    \caption{Training pipeline of our proposed PiDR framework.}
    \label{fig:PiDR}
\end{figure*}

\noindent We propose PiDR, a physics-informed inertial dead-reckoning framework for autonomous platforms in situations of pure inertial navigation. It fuses the underlying physical constraints of the strapdown
INS equations of motion with sensor data to estimate the position, velocity, and orientation of autonomous platforms. To this end, a composite loss function is derived as a weighted combination of physics-based loss and supervised data loss to emulate the principles of inertial navigation. During training, the physics-based constraints act as soft regularizes, ensuring that the learned trajectories remain physically plausible. This formulation enables operation under limited or sparse GT availability while improving generalization across unseen trajectories and sensor configurations.  
Our PiDR framework is illustrated in Fig. \ref{fig:PiDR}. The arrow colors distinguish between the experimental data flow (blue) and the generated collocation points (red). The neural network takes the platform’s inertial measurements (specific force and angular velocity) along with the synchronized time stamps as input and predicts the navigation states (position, velocity, and orientation). In the following section, we elaborate on each part of PiDR.

\subsection{Physics-Informed Neural Network for Inertial Dead Reckoning}

\noindent To estimate position, velocity, and orientation using IMU measurements, we offer a PiDR model.  It embeds the physics of strapdown INS  equations of motion in \eqref{eq:pos}, \eqref{eq:vel} and \eqref{eq:ori} into a neural network training pipeline.

\noindent We employ a fully-connected feedforward neural network with $L$ hidden layers, each containing $H$ neurons and nonlinear activation functions $\sigma(\cdot)$. Let $\mathcal{N}_{\hat{\vartheta}}$
denote the network with trainable parameters $\hat{\vartheta}$ that approximates the nonlinear mapping

\begin{equation}
	\begin{split}
		\mathcal{N}_{\hat{\vartheta}}(\mathbf{u}) : & (t,\; f_x,\; f_y,\; f_z,\; \omega_x,\; \omega_y,\; \omega_z)
		\;\longrightarrow\; \\
		& (\hat{x}, \hat{y}, \hat{z}, \hat{v}{_x}, \hat{v}{_y}, \hat{v}{_z}, \hat{\phi}, \hat{\theta}, \hat{\psi}),
	\end{split}
\end{equation}
such that:
\begin{equation}
	\mathcal{N}_{\hat{\vartheta}}(\mathbf{u})
	=
	\mathbf{W}_L \,\sigma\Big(
	\cdots \sigma\big( \mathbf{W}_2 \, \sigma(\mathbf{W}_1 \mathbf{u} + \mathbf{b}_1) + \mathbf{b}_2 \big)
	\cdots
	\Big) + \mathbf{b}_L ,
\end{equation}
where $\mathbf{u} \in \mathbb{R}^7$ is the input vector of the time step and inertial measurements, $\mathcal{N}_{\hat{\vartheta}}(\mathbf{u}) \in \mathbb{R}^9$ is the network output of the navigation solution (viz., the position, velocity, and orientation) and $\mathbf{W}_\ell$, $\mathbf{b}_\ell$ are the trainable weights and biases, respectively.

\noindent During the training process, a composite loss function, defined to this end, is minimized to obtain the optimized trainable parameters. The individual loss components are described in the subsequent subsections.

\vspace{0.2cm}
\subsubsection{Data-driven Loss Component}

The data-driven component of the objective function enforces agreement between the predicted and GT states. 
Let the GT vectors  denoted by $\mathbf{p}, \mathbf{v}$, and $\boldsymbol{\eta}$. The corresponding predicted output be $ \mathcal{N}_{\hat{\vartheta}}(\mathbf{u}) = [\hat{\mathbf{p}}, \hat{\mathbf{v}}, \hat{\boldsymbol{\eta}}]$, where 
$\hat{\mathbf{p}} = [\hat{x}, \hat{y}, \hat{z}]$, 
$\hat{\mathbf{v}} = [\hat{v}{_x}, \hat{v}{_y}, \hat{v}{_z}]$, and 
$\hat{\boldsymbol{\eta}} = [\hat{\phi}, \hat{\theta}, \hat{\psi}]$. 
The data loss is defined as a weighted mean squared error (MSE):

\begin{equation}
	\begin{split}
		\mathcal{L}_{\text{data}} 
		&=
		\frac{1}{N} \sum_{i=1}^{N} \Big(
		w_p \, \big\|\hat{\mathbf{p}}^{(i)} - \mathbf{p}^{(i)}\big\|_2^2
		+
		w_v \, \big\|\hat{\mathbf{v}}^{(i)} - \mathbf{v}^{(i)}\big\|_2^2 \\ 
		&\quad
		+
		w_\eta \, \big\|\hat{\boldsymbol{\eta}}^{(i)} - \boldsymbol{\eta}^{(i)}\big\|_2^2
		\Big).
	\end{split}
	\label{eq:dr_data_loss_euler}
\end{equation}
where \(w_p\), \(w_v\), and \(w_{\eta}\) are weighting coefficients that balance the relative contributions of position, velocity, and orientation estimation errors, respectively. 
Additionally, $N$ denotes the number of synchronized training samples at which both network-predicted inertial data 
($\hat{\mathbf{p}}(t_i)$, $\hat{\mathbf{v}}(t_i)$, $\hat{\boldsymbol{\eta}}(t_i)$) and the corresponding GT measurements $(\mathbf{p}(t_i)$, $\mathbf{v}(t_i)$, $\boldsymbol{\eta}(t_i))$ are available. The time instances $t_i$ correspond to the timestamps of the GT data after alignment with the high-frequency IMU measurements, and $\|\cdot\|$ denotes the Euclidean norm. In our experiment, we fix $w_p=w_v=w_\eta=1$.

\vspace{0.2cm}

\subsubsection{Physics-informed Loss Component}
\noindent The physics-informed component learns the strapdown inertial navigation states
\(
\big\{\hat{\mathbf{p}}^n(t_i),\; \hat{\mathbf{v}}^n(t_i),\; \hat{\mathbf{C}}_b^n(t_i)\big\}
\)
from the IMU-measured specific force and angular velocity vectors $\left\{ \mathbf{f}_{ib}^b,\; \boldsymbol{\omega}_{ib}^b \right\}$
at time $t_i$ by enforcing the underlying kinematic and dynamic equations. 
Let \(\mathbf{r}_{\text{phys}}\) is the residual obtained by substituting the neural network outputs of $\mathcal{N}_{\hat{\vartheta}}(\mathbf{u})$ and defined as:
\begin{equation}
	\mathbf{r}_{\text{phys}} =
	\mathcal{F}\!\left(
	\mathcal{N}_{\hat{\vartheta}},\,
	\frac{d}{dt}\mathcal{N}_{\hat{\vartheta}},\,
	\mathbf{f},\,
	\boldsymbol{\omega}
	\right),
\end{equation}
where $\mathcal{F}(\cdot)$ represents the strapdown INS kinematics and dynamics equations.  Enforcing $\mathbf{r}_{\text{phys}} < \epsilon$
ensures that the learned trajectory adhere to inertial dynamics.

\noindent The strapdown INS equations impose three sets of constraints:

\noindent\textbf{Position:}\quad
\begin{equation}
	\mathbf{r}_p
	= 
	\frac{d\hat{\mathbf{p}}^n}{dt}
	-
	\mathbf{D}(\varphi,h)\,\hat{\mathbf{v}}^n,
	\label{eq:res_pos}
\end{equation}
where $\mathbf{D}(\varphi,h)$ is defined in~\eqref{eq:pos}.

\noindent\textbf{Velocity:}\quad
\begin{equation}
	\mathbf{r}_v
	=
	\frac{d\hat{\mathbf{v}}^n}{dt}
	-
	\left(
	\hat{\mathbf{C}}_b^n\,\mathbf{f}_{ib}^b
	-
	\big(2\boldsymbol{\omega}_{ie}^n + \boldsymbol{\omega}_{en}^n\big)
	\times
	\hat{\mathbf{v}}^n
	+
	\mathbf{g}^n
	\right).
	\label{eq:res_vel}
\end{equation}

\noindent\textbf{Orientation:}\quad
\begin{equation}
	\mathbf{r}_\eta
	=
	\frac{d\hat{\mathbf{C}}_b^n}{dt}
	-
	\left(
	\hat{\mathbf{C}}_b^n\,\boldsymbol{\Omega}_{ib}^b
	-
	\big(\boldsymbol{\Omega}_{ie}^n + \boldsymbol{\Omega}_{en}^n\big)\,\hat{\mathbf{C}}_b^n
	\right).
	\label{eq:res_att}
\end{equation}

\noindent The overall physics residual vector $\mathbf{r}_{\text{phys}}$, is constructed by stacking the individual residual components corresponding to the inertial navigation equations. Specifically, it is defined as
\begin{equation}
	\mathbf{r}_{\text{phys}}
	=
	\begin{bmatrix}
		\mathbf{r}_p^\top &
		\mathbf{r}_v^\top &
		\mathbf{r}_\eta^\top
	\end{bmatrix}^{\!\top}.
	\label{eq:pinn}
\end{equation}

\noindent  Accordingly, the physics-informed loss is formulated as the MSE of these residuals over \(N_{p}\) collocation samples:
\begin{equation} \label{eq:phy}
	\mathcal{L}_{\text{phys}}
	=
	\frac{1}{N_p}
	\sum_{i=1}^{N}
	\Big(
	\big\|\mathbf{r}_p^{(i)}\big\|_2^2
	+
	\big\|\mathbf{r}_v^{(i)}\big\|_2^2
	+
	\big\|\mathbf{r}_\eta^{(i)}\big\|_2^2
	\Big)
	=
	\frac{1}{N}
	\sum_{i=1}^{N}
	\big\|
	\mathbf{r}_{\text{phys}}^{(i)}
	\big\|_2^2 .
\end{equation}
This loss penalizes deviations from the physical motion laws, thus constraining the neural network to learn dynamics consistent with inertial navigation equations. Furthermore, due to the use of low-cost inertial sensors and the relatively short time experiments, both the Earth's rotation rate and transport rate are neglected in 
{(\ref{eq:res_vel})} and \eqref{eq:res_att}.

\vspace{0.2cm}
\subsubsection{Total Objective Function}

\noindent The overall training objective of the PiDR jointly enforces agreement with measured data and physical consistency. The optimal network parameters are obtained by solving the following minimization problem
\begin{equation}
	\hat{\theta}
	=
	\arg\min_{\vartheta}\;
	\mathcal{L}_{\text{total}}(\vartheta),
\end{equation}
where
\begin{equation}
	\mathcal{L}_{\text{total}} =
	\lambda_{\text{data}} \,	\mathcal{L}_{\text{data}} +
	\lambda_{\text{phys}} \, \mathcal{L}_{\text{phys}},
	\label{eq:dr_total_loss}
\end{equation}
In \eqref{eq:dr_total_loss}, $\lambda_{\text{data}}$ is a hyperparameter that balances the relative contribution of the data-driven loss, and $\lambda_{\text{phys}}$ is a hyperparameter controlling the relative importance of the physics-informed loss. 
In our experiment, we fix $\lambda_{\text{data}}=3$ and $\lambda_{\text{phys}}=2$.

\vspace{0.1cm}
\subsection{PiDR Architecture and Implementation}
\label{subsec:pidr_arch}

\noindent
The PiDR model is implemented as a fully connected feedforward neural network designed for inertial navigation. The architecture follows a fixed-depth, fixed-width design to ensure stable optimization over long trajectories while maintaining sufficient expressive capacity.

\noindent
The network consists of an input layer, followed by four hidden layers with a uniform width of 128 neurons per layer, and a linear output layer. Each hidden layer is equipped with a ReLU activation function. ReLU activation function \cite{fukushima2007visual} is adopted for its numerical stability and improved gradient flow properties in DNN. In order to reduce overfitting and improve generalization across numerous sensor configurations and trajectories, dropout regularization with a rate of 0.1 is applied after each hidden layer. AdamW optimizer~\cite{loshchilov2017decoupled} with an initial learning rate of $10^{-3}$
is employed using \texttt{PyTorch}. The PiDR training algorithm is presented in Algorithm 1.

\begin{algorithm}[!t]
\caption{PiDR Training Procedure for Strapdown Inertial Navigation}
\label{alg:pidr_training}

Initialize network parameters $\Theta=(\mathbf W_0,\mathbf b_0)$\;

Set iteration counter $k\leftarrow0$ and threshold $\varepsilon$\;

\For{each epoch}{

    \For{each mini-batch trajectory $m$}{

        Predict
        $\hat{\mathbf x}^{(m)}(t)=
        \mathcal N_{\hat\vartheta}
        (t,\mathbf f,\boldsymbol\omega)$\;

        Compute $\mathcal L_{\rm data}$
        using (\ref{eq:dr_data_loss_euler})\;

        Generate $N_p$ collocation points\;

        Compute $\mathcal L_{\rm phys}$
        using (\ref{eq:phy})\;

        Compute total loss
        $\mathcal L_{\rm total}$
        using (\ref{eq:dr_total_loss})\;

        Update $\Theta$ using backpropagation\;

        \If{$\mathcal L_{\rm total}<\varepsilon$}{
            \Return{$\Theta$}
        }

        $k\leftarrow k+1$\;
    }
}

\Return{$\Theta$}

\end{algorithm}














\noindent
To ensure robustness across varying motion patterns, PiDR is trained simultaneously on multiple trajectories. The physics collocation points ($N_p$) are drawn uniformly from all trajectories, thereby enforcing consistency over the entire temporal domain rather than only at supervised data points.
Tables \ref{tab:gpu_config} and~\ref{tab:pidr_params} report the configuration of the hardware and software parameters, respectively.

\begin{table}[!ht]
	\centering
	\caption{Hardware configuration for PiDR training.}
	\label{tab:gpu_config}
    \setlength{\tabcolsep}{15pt}
	\renewcommand{\arraystretch}{0.85}
	\begin{tabular}{ll}
		\toprule
		\textbf{Component} & \textbf{Specification} \\
		\midrule
		GPU Model             & NVIDIA GeForce RTX 4090 \\
		
		GPU OS            & Linux (Debian) \\
		Architecture      & x86\_64 \\
		
		CUDA Version      & 11.8 \\
		cuDNN Version     & 9.1 \\
		System RAM        & 67.26 GB \\
		GPU Memory        & 25.28 GB \\
		Tensor Cores             & 512\\
		CPU Cores         & 24 \\
		
		\bottomrule
	\end{tabular}
\end{table}

\begin{table}[!ht]
	\centering
	\caption{Software configuration for PiDR training.}
	\label{tab:pidr_params}
    \setlength{\tabcolsep}{15pt}
	\renewcommand{\arraystretch}{0.85}
	\begin{tabular}{ll}
		\toprule
		\textbf{Parameter} & \textbf{Value} \\
		\midrule
		Python Version    & 3.9.21 \\
		PyTorch Version   & 2.5.1 \\
		Hidden layers                & 4 \\
		Neurons per layer            & 128 \\
		Activation function          & ReLU \\
		Dropout rate                 & 0.1 \\	
		
		Optimizer                    & AdamW \\
		Initial learning rate        & $1 \times 10^{-3}$ \\
		Weight decay                 & $1 \times 10^{-5}$ \\
		
		Scheduler factor             & 0.1 \\
		
		Batch size                   & 512 \\
		
		Gradient clipping            & $\ell_2$ norm, max 1.0 \\
		Collocation points           & 1400  \\
        Training time       & 411.20 s\\
        Peak GPU memory     & 1.25 GB\\
		
		\bottomrule
	\end{tabular}
\end{table}

\subsection{PiDR-EKF}
\noindent To further enhance estimation accuracy, an EKF is integrated with the proposed PiDR framework. 
The hybrid PiDR--EKF model operates on a 15-dimensional state vector

\begin{equation}
\mathbf{x}_k = 
\begin{bmatrix}
\mathbf{p}_k^\top & 
\mathbf{v}_k^\top & 
\boldsymbol{\eta}_k^\top & 
\mathbf{b}_{g,k}^\top & 
\mathbf{b}_{a,k}^\top
\end{bmatrix}^\top,
\end{equation}
where $\mathbf{p}_k \in \mathbb{R}^3$, $\mathbf{v}_k \in \mathbb{R}^3$, and $\boldsymbol{\eta}_k \in \mathbb{R}^3$ denote position, velocity, and orientations, while $\mathbf{b}_{g,k}$ and $\mathbf{b}_{a,k}$ represent gyroscope and accelerometer biases, respectively.

\noindent While the complete EKF formulation is available in standard navigation literature~\cite{titterton2004strapdown, farrell2008aided, groves2013book}, this section defines only the PiDR-based measurement model.

\noindent To this end, the PiDR outputs, namely the position, velocity, and orientation are used as external measurements. The PiDR measurement is:

\begin{equation}
\mathbf{z}_k^{\text{PiDR}} = 
\begin{bmatrix}
(\mathbf{p}_k^{\text{PiDR}})^\top & 
(\mathbf{v}_k^{\text{PiDR}})^\top & 
(\boldsymbol{\eta}_k^{\text{PiDR}})^\top
\end{bmatrix}^\top
\end{equation}

The corresponding measurement matrix is given by
\begin{equation}
\mathbf{H}^{\mathrm{PiDR}} =
\begin{bmatrix}
\mathbf{I}_3 & \mathbf{0}_{3\times3} & \mathbf{0}_{3\times3} & \mathbf{0}_{3\times3} & \mathbf{0}_{3\times3} \\
\mathbf{0}_{3\times3} & \mathbf{I}_3 & \mathbf{0}_{3\times3} & \mathbf{0}_{3\times3} & \mathbf{0}_{3\times3} \\
\mathbf{0}_{3\times3} & \mathbf{0}_{3\times3} & \mathbf{H}_{\eta} & \mathbf{0}_{3\times3} & \mathbf{0}_{3\times3}
\end{bmatrix},
\end{equation}
where $\mathbf{H}_{\eta}$ maps the attitude misalignment error states to the Euler angle errors, and is given by
\begin{equation}
\mathbf{H}_{\eta} =
\begin{bmatrix}
\dfrac{\cos \psi}{\cos \theta} & \dfrac{\sin \psi}{\cos \theta} & 0 \\
-\sin \psi & \cos \psi & 0 \\
\tan \theta \cos \psi & \tan \theta \sin \psi & 1
\end{bmatrix}.
\end{equation}

\noindent While a correlation exists between the process and measurement noise covariances due to their shared reliance on inertial data, it is neglected here as the readings are subject to different nonlinear modeling. To compensate, a conservative inflation of the measurement noise covariance is employed, ensuring the filter does not become over-reliant on PiDR outputs.


\section{Analysis and Experimental Resutls} \label{sec: analysis}

\noindent We begin this section by describing both the datasets used in our experiment, followed
by definitions of evaluation metrics. Then, we compare the performance of PiDR and PiDR-EKF approaches against the baselines.

\subsection{Inertial Datasets}
\label{sec:datasets}
\noindent
To evaluate the robustness and generalization capability of our proposed framework on multiple platforms, we carried out experiments on datasets collected using: 
(i) a wheeled mobile robot and
(ii) an AUV.
Each dataset provides raw inertial sensor measurements and the corresponding GT trajectories.
The two datasets differ substantially in terms of platform dynamics, operating environments, sensor configurations,  and navigation constraints, thereby enabling a comprehensive validation of the proposed approach.

\subsubsection{Mobile robot dataset}
\noindent
The dataset was collected using a Husarion \emph{ROSbot XL}~\cite{husarion_ROSbot_xl} autonomous platform at the University of Haifa, Israel, parking lot \cite{yampolsky2024multiple}.  \emph{ROSbot XL} is a wheeled robot of dimensions 332 [mm] $\times$ 325 [mm] $\times$ 133.5 [mm]. 
The platform was equipped with nine IMUs, out of which data generated by three distinct IMUs (one for training and two for testing) were used in this research. These IMUs are DOT IMUs manufactured by Xsens Technologies \cite{xsens_dot}. The specifications of IMU are provided in Table \ref{tab:xsens_spec}.

\begin{table}[!ht]
	\centering
	\caption{Xsens DOT IMU sensor specifications~\cite{xsens_dot}.}
	\renewcommand{\arraystretch}{1.15}
	
	\begin{tabular}{lcc}
		\toprule
		\textbf{Specification} 
		& \textbf{Accelerometer} 
		& \textbf{Gyroscope} \\
		\midrule
		Sampling Rate [Hz] 
		& 120 
		& 120 \\
		
		Bias In-run Stability 
		& 0.03\,mg 
		& 10\,°/h \\
		
		Noise Density 
		& 120\,mg/$\sqrt{\text{Hz}}$ 
		& 0.007\,°/s/$\sqrt{\text{Hz}}$ \\
		\bottomrule
	\end{tabular}
	\label{tab:xsens_spec}
\end{table}

\begin{figure*}[!t]
	\centering
	\includegraphics[
	width=0.85\textwidth,
	height=0.35\textheight,
	keepaspectratio
	]{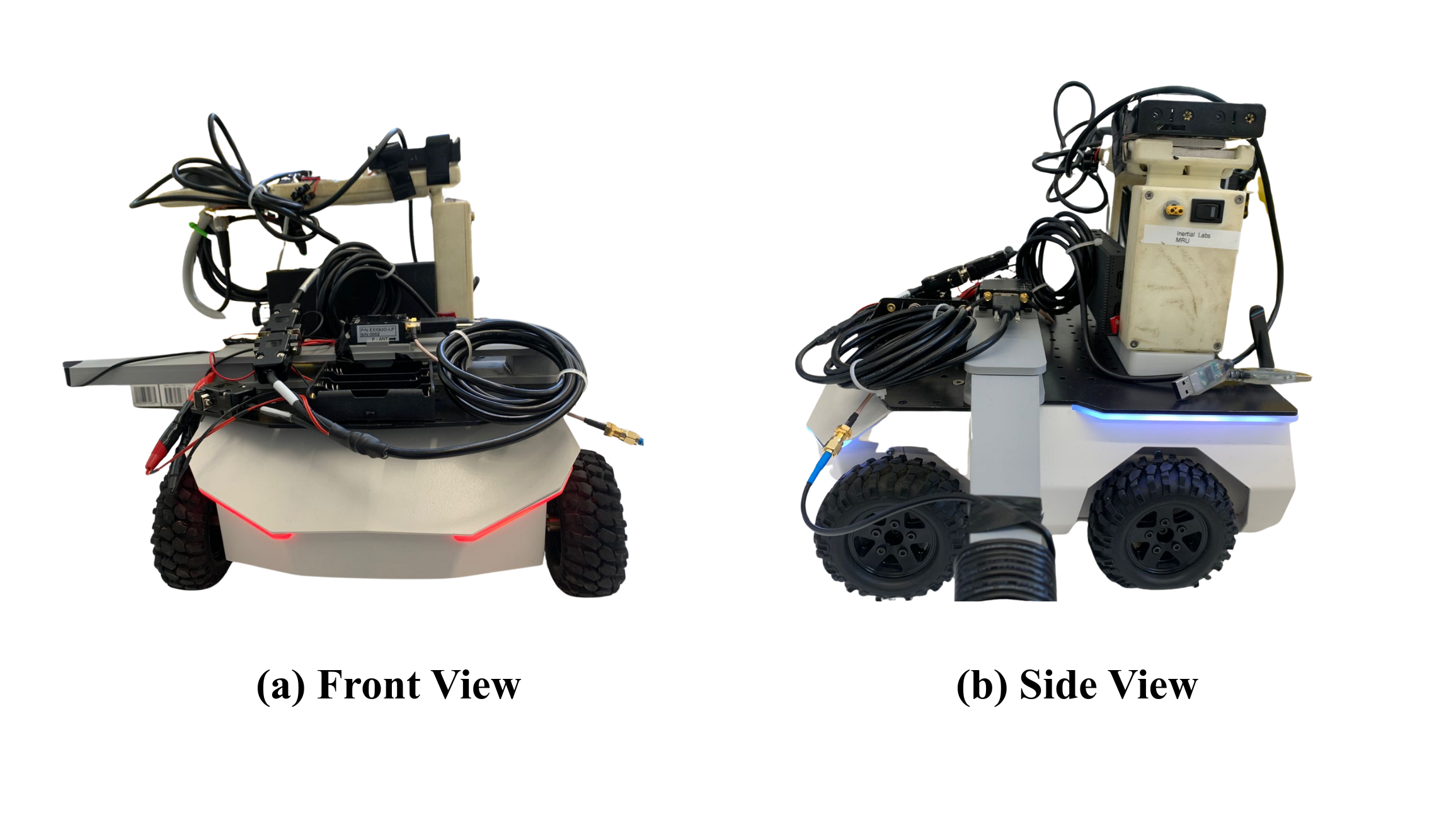}
	\caption{ROSbot XL mounted with DOT IMUs and MRU-P.}
	\label{fig:ROSbot}
\end{figure*}

\noindent
The GT trajectories are derived from an MRU-P~\cite{inertiallabs_mrup_2023} equipped with a licensed GNSS real-time kinematic (RTK) TerraStar-C Pro system ~\cite{terrastar_cpro}. As illustrated in Fig.~\ref{fig:ROSbot}, the DOT IMUs and MRU-P sensor are rigidly mounted at distinct locations on the robot body, resulting in different sensor placement configurations.

\begin{figure} [!ht]
		\begin{center}
			\includegraphics[height=6.cm,width=9cm]{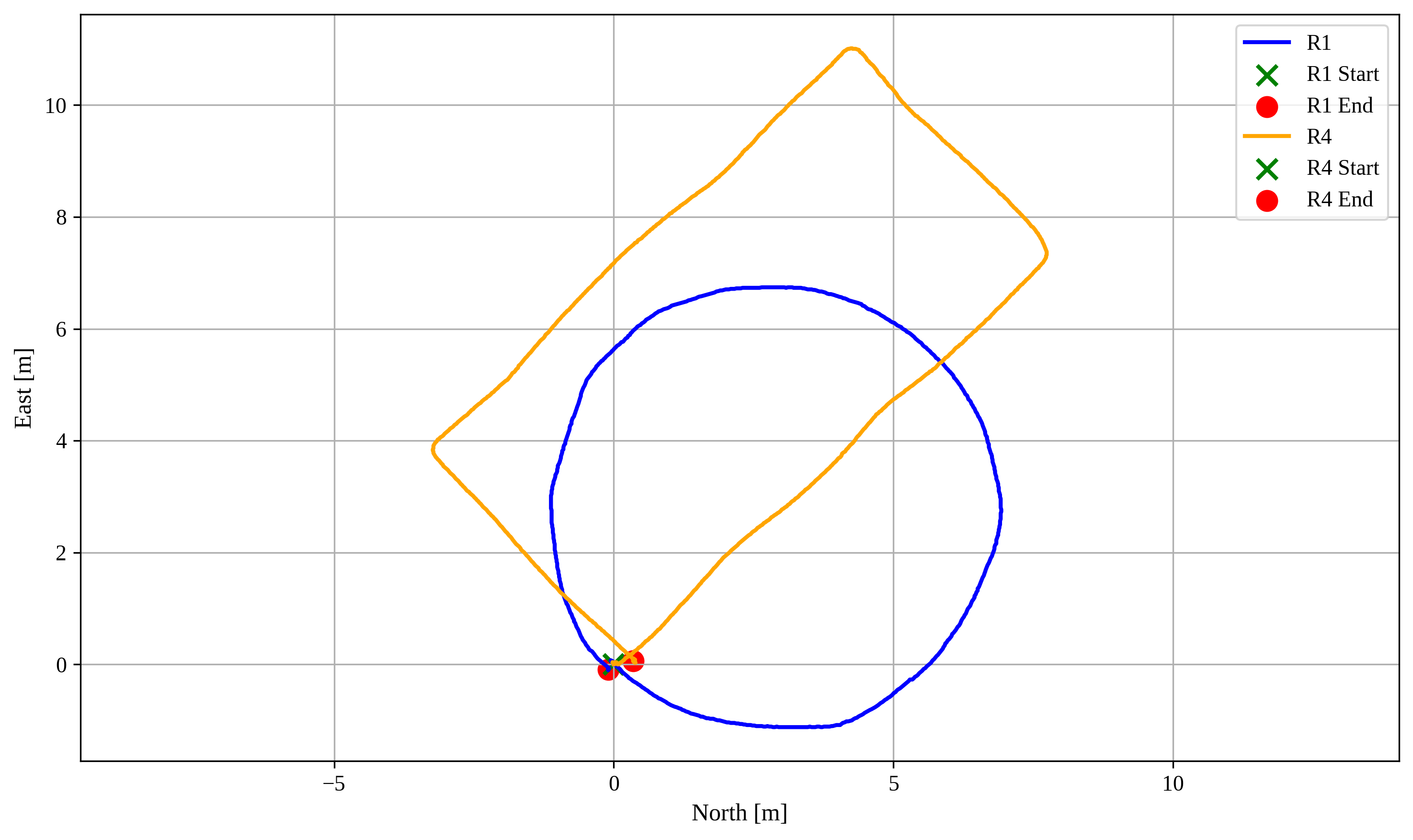}
			\caption{ROSbot XL trajectories used for training of PiDR (Trajectories R1 and R4).}
			\label{fig:traj_ros_train}
		\end{center}
	\end{figure}

\noindent To evaluate the PiDR model, training and testing are conducted using trajectories recorded by different IMUs.
The model has been trained on trajectories R1 and R4 (Fig. \ref{fig:traj_ros_train}) with a duration of 11 [min].
Subsequently, the model was tested on unseen circular trajectories (R2 and R3) and rectangular trajectories (R5 and R6) with a duration of 20 [min], using different IMUs (one for training and two for testing) mounted at different locations on the robot. The cumulative path lengths of the training and testing sets are 76 [m] and 154 [m], respectively.

\subsubsection{AUV dataset}
\noindent
The second dataset is collected using Snapir AUV in the Mediterranean Sea near Haifa, Israel \cite{shurin2022autonomous}. Snapir is an ECA robotics, a modified A18D mid-size AUV designed for deep-water applications up to 3000 [m] depth with 21 hours of endurance~\cite{eca_a18d_2023}. 

\begin{figure} [!ht]
	\begin{center}
		\includegraphics[height=6cm,width=9cm]{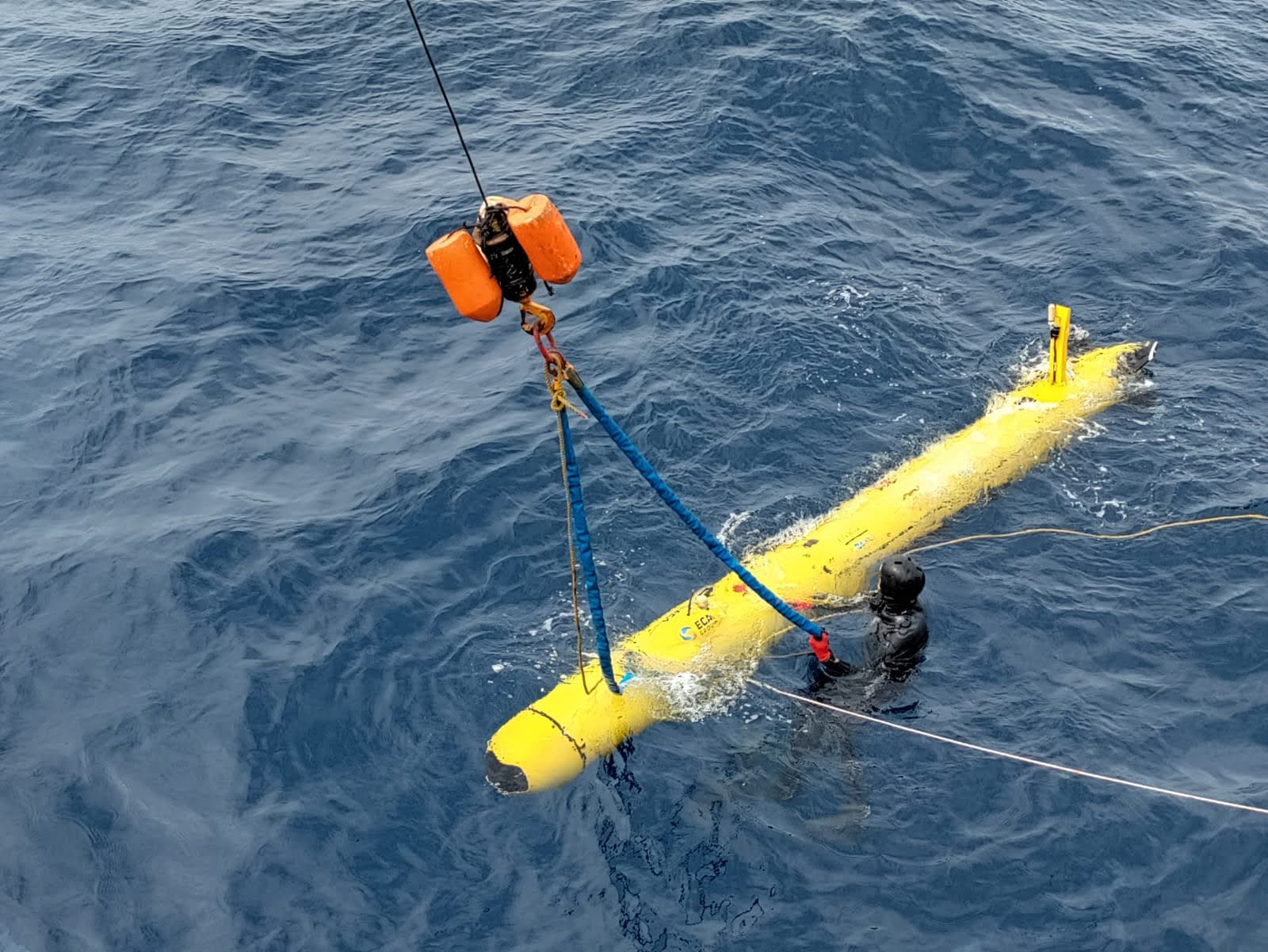}
		\caption{Snapir AUV during the mission in the Mediterranean Sea, Haifa, Israel.}
		\label{fig:auv}
	\end{center}
\end{figure}

\noindent
Snapir AUV is equipped with various MEMS sensors for DR evaluation in underwater environments. 
The Snapir, as shown in Fig. \ref{fig:auv}, is equipped with (a) iXblue Phins Subsea, which is a fibre-optic gyroscope-based, high-performance subsea INS~\cite{ixblue_phins_2023}, and (b) Teledyne RDI Work Horse navigator  DVL~\cite{teledyne_dvl_2023} that achieve accurate velocity measurements with a standard deviation of 0.02 [m/s]. The INS samples at 100 [Hz], while the DVL samples at 1 [Hz].

\begin{figure} [!ht]
		\begin{center}
			\includegraphics[height=9cm,width=9cm]{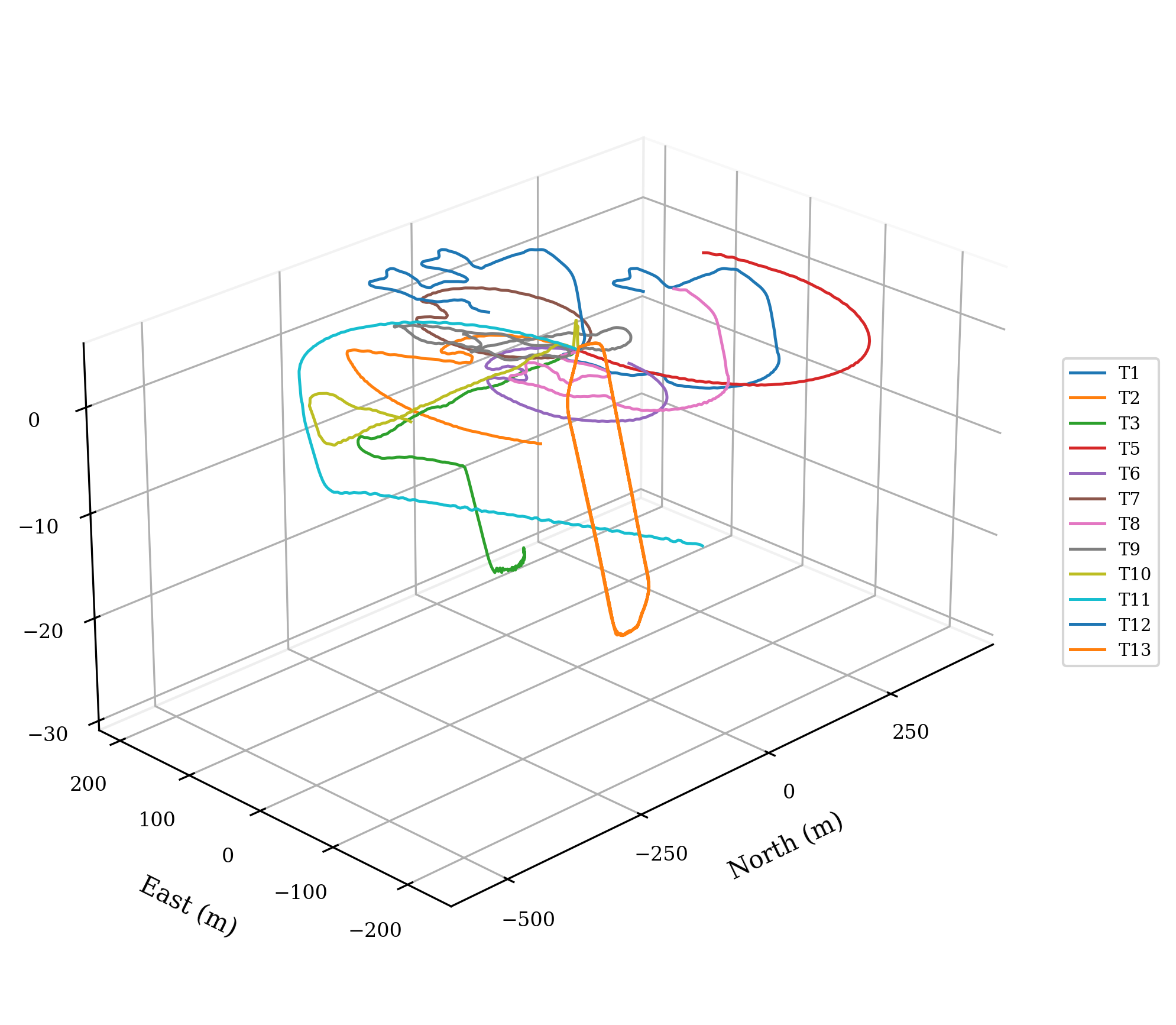}
			\caption{AUV trajectories in the NED frame.}
			\label{fig:traj_auv_train}
		\end{center}
	\end{figure}

\noindent
Snapir AUV performs diverse maneuvers in underwater, each of a duration of 6.6 [min]. All these missions vary in trajectory shape, depth, and speed as presented in Fig.~\ref{fig:traj_auv_train}. 
The primary training set consists of trajectories T1, T2, T3, T5, T6, T7, and T8, comprising approximately 53.3~[min] of data. To evaluate generalization under previously unseen scenarios, the test set includes trajectories T10, T11, and T13, with a total duration of approximately 20~[min]. The cumulative path lengths of the training and testing sets are $2000\,\mathrm{m}$ and $983\,\mathrm{m}$, respectively. In addition, two further cross-validations were conducted on the AUV dataset using alternative train--test splits. However, for clarity and conciseness, the quantitative analysis presented in this study is limited to the primary experimental setup only.
The GT is provided by post-processing software Delph INS~\cite{ixblue_delph_2023} for INS-based subsea navigation.

\subsubsection{Summary} 
\noindent Table \ref{tab:dataset_summary} summarizes the dataset key parameters, including the cross-validation datasets. The overall dataset spans 229 [min], of which 129 [min] are used for training and 100 [min] for testing. In the primary study, a subset totaling 97 [min] is considered, comprising 56 [min] for training and 41 [min] for testing, while the remaining data are reserved for cross-validation experiments.
 That is a train/test ratio of (57\%/43\%) instead of the common practice of (80\%/20\%) resulting in more than twice data for testing.

\begin{table}[!ht]
	\centering
	\caption{Main dataset parameters.}
	\label{tab:dataset_summary}
    \setlength{\tabcolsep}{10pt}
	\renewcommand{\arraystretch}{0.75}
	\begin{tabular}{lcc}
		\toprule
		\textbf{Attribute} & \textbf{ROSbot XL} & \textbf{Snapir AUV } \\
		\toprule
		Sampling rate (IMU) & 120 [Hz] & 100 [Hz] \\
		Sampling rate (GT) & 5 [Hz] & 1 [Hz] \\
		
		Train trajectories & 2 & 21 (3 sets in total) \\
		Test trajectories & 4 & 9 (3 sets in total) \\
		Each trajectory duration & 4-6 [min] & 6.6 [min]\\
		
		Total duration & 31 [min] & 198 [min] \\

		\bottomrule
	\end{tabular}
\end{table}

\subsection{Evaluation Metrics} \label{metric}

\noindent To evaluate trajectory estimation performance, we employ the following evaluation metrics.
Let the predicted and GT positions at time $t_i$ be denoted as $\hat{\mathbf{p}}(t_{i})$ and $\mathbf{p}(t_i)$, respectively.

\noindent\textbf{Absolute Trajectory Error (ATE):}  
\begin{equation}
	\begin{split}
		&\text{ATE}_i
		=
		\left\|
		\hat{\mathbf{p}}(t_{i})-\mathbf{p}(t_i)
		\right\|_2.
	\end{split}
\end{equation}

\noindent\textbf{Position Root Mean Square Error (PRMSE):}  
\begin{equation}
	\text{PRMSE}
	=
	\sqrt{\frac{1}{N}
		\sum_{i=1}^{N} (\text{ATE}_i)^2 },
\end{equation}
where $N$ is the total number of samples.

\noindent\textbf{Total Distance Error (TDE):}  
\begin{equation}
	\text{TDE} (\%)
	=
	\frac{\text{PRMSE}}{D}
	\times
	100.
\end{equation}
where $D$ is the distance of trajectory.

\noindent\textbf{Final Distance Error (FDE):} 
\begin{equation}
	\text{FDE} = \left\| \hat{\mathbf{p}}(t_{\text{end}}) - \mathbf{p}(t_{\text{end}}) \right\|_2,
\end{equation}
where 
$\hat{\mathbf{p}}(t_{\mathrm{end}})$ is the predicted final position,
$\mathbf{p}(t_{\mathrm{end}})$ is the GT final position.

\subsection{Performance Analysis}
\noindent This subsection presents a comparison between the proposed PiDR framework and baselines. 
The comparison has been done using both local accuracy (PRMSE, MATE) and global trajectory consistency (TDE, FDE) described in Section~\ref{metric}.

\subsubsection {Baseline approaches} 
\noindent We compare our approach against three other methods: 1) INS - the model-based commonly used inertial equations of motion \eqref{eq:pos}, \eqref{eq:vel}, and \eqref{eq:ori},  where a 2D implementation is used for the mobile robot and a 3D for the AUV, 2)  The model-based MoRPI approach \cite{etzion2023morpi} originally designed to handle mobile robots moving in periodic trajectories, and 3) MoRPI-PINN~\cite{sahoo2025morpi} its physics-informed counterpart.

\subsubsection{Mobile Robot}
\begin{figure*}[!ht]
	\centering
	
	\subfloat[\normalfont\rmfamily Circular trajectories (R2 and R3)]{
		\includegraphics[width=0.45\textwidth]{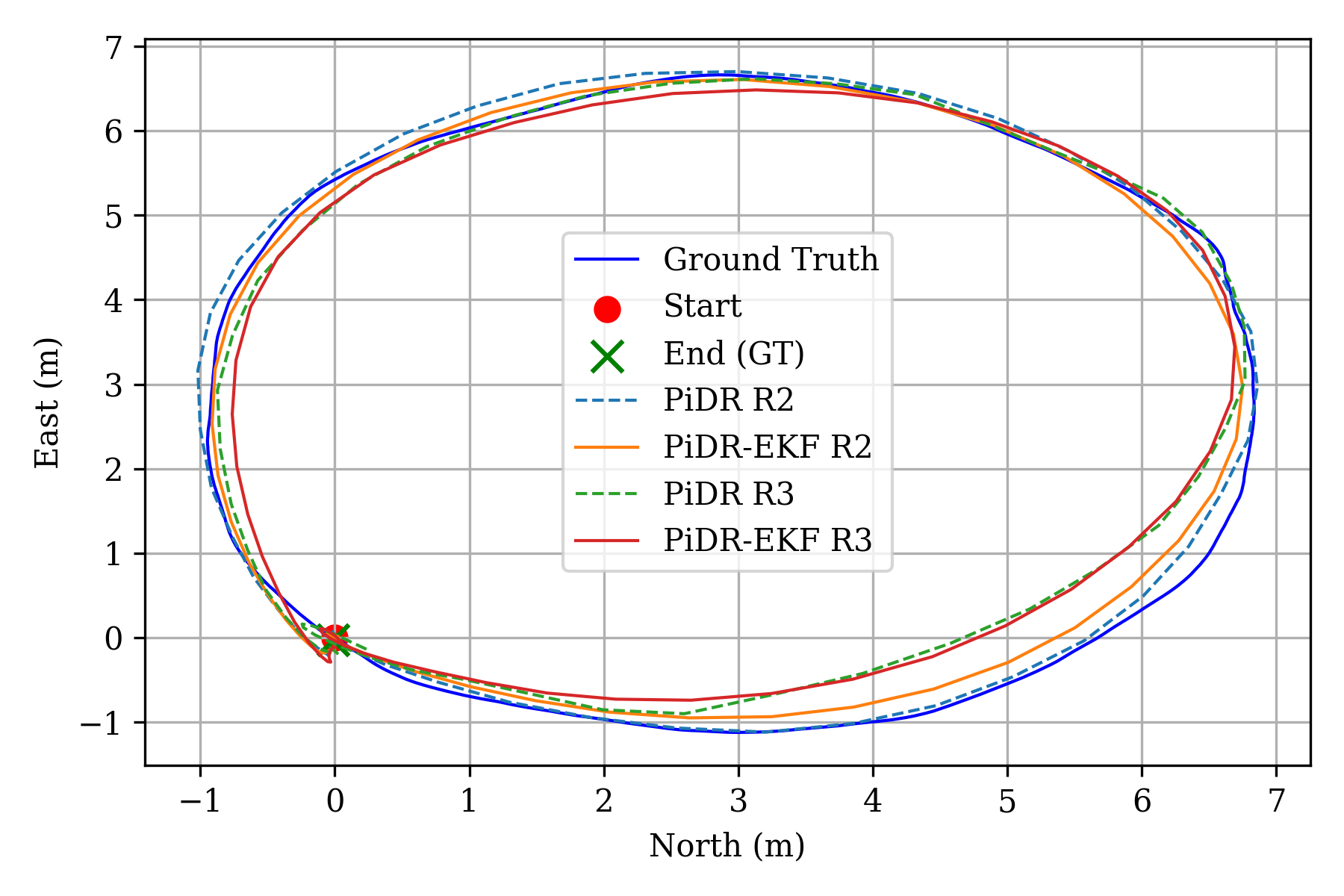}
	}\hfill
	\subfloat[\normalfont\rmfamily Rectangular trajectories (R5 and R6)]{
		\includegraphics[width=0.45\textwidth]{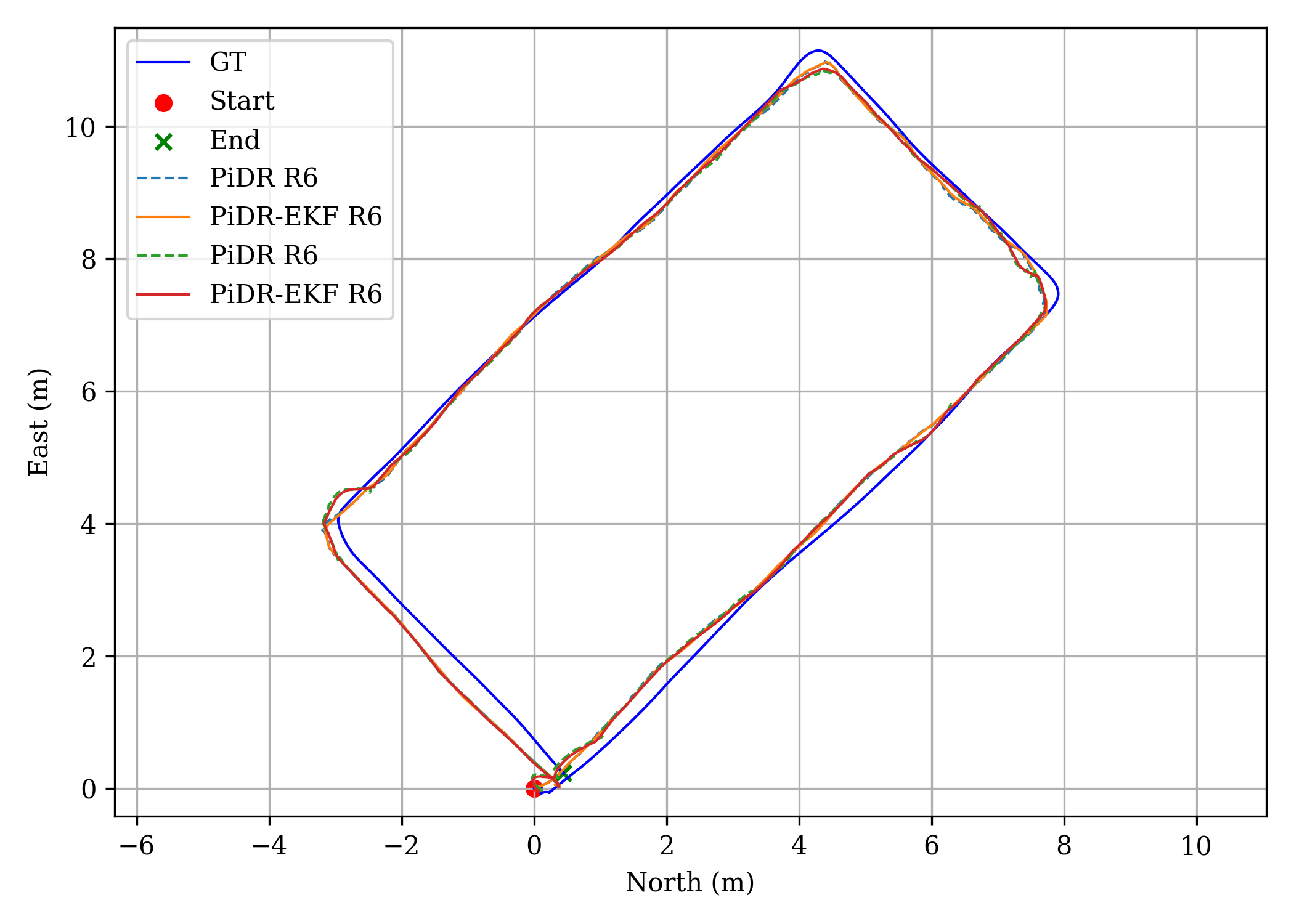}
	}
	
	\caption{Comparison of GT, PiDR, and PiDR-EKF estimated trajectories (N--E) for the mobile robot.}
	\label{fig:traj_comparison_ros}
\end{figure*}

\noindent The proposed PiDR approach underwent evaluation for mobile robot using test set
trajectories R2, R3, R5, and R6. Fig. \ref{fig:traj_comparison_ros} illustrates the position comparison for GT and PiDR models in the n-frame for test
trajectories.

\begin{table*}[!ht]
	\centering
	\caption{Evaluation of the proposed PiDR and PiDR-EKF on the mobile robot trajectories.}
	\label{tab:ROSbot}
	\renewcommand{\arraystretch}{0.85}
	
	\begin{tabularx}{\textwidth}{l l
			>{\centering\arraybackslash}X
			>{\centering\arraybackslash}X
			>{\centering\arraybackslash}X
			>{\centering\arraybackslash}X
			>{\centering\arraybackslash}X
			>{\centering\arraybackslash}X}
		
		\toprule
		\multirow{2}{*}{\textbf{Metric}} 
		& \multirow{2}{*}{\textbf{Method}}
		& \multicolumn{2}{c}{\textbf{Circular}}
		& \multicolumn{2}{c}{\textbf{Rectangular}}
		& \multirow{2}{*}{\textbf{Average}}
		& \multirow{2}{*}{%
			\parbox{1.9cm}{%
				\centering
				\textbf{Improvement [\%] using PiDR-EKF}
		}}\\
		
		\\[-0.6ex]
		\cmidrule(lr){3-4} \cmidrule(lr){5-6}
		
		& & \textbf{R2} & \textbf{R3}
		& \textbf{R5} & \textbf{R6}
		& & \\
		
		\midrule
		
		\multirow{5}{*}{\textbf{PRMSE [m]}}
		& 2D INS
		& 6.7 & 7.8 & 8.3 & 9.6
		& 8.1 & 68 \\
		
		& MoRPI
		& 16.2 & 16.2 & 23.5 & 23.1
		& 18.9 & 86 \\
		
		& MoRPI-PINN
		& 5.0 & 4.6 & 1.3 & 0.9
		& 2.9 & 11 \\
		
		& PiDR (ours)
		& 5.0 & 4.5 & 0.3 & 0.4
		& 2.6 & 0.2 \\
		
		& PiDR-EKF (ours)
		& 5.0 & 4.5 & 0.3 & 0.4
		& 2.6 & -- \\
		
		\midrule
		
		\multirow{5}{*}{\textbf{MATE [m]}}
		& 2D INS
		& 6.2 & 7.5 & 7.8 & 8.7
		& 7.6 & 73 \\
		
		& MoRPI
		& 13.2 & 13.2 & 20.0 & 19.5
		& 15.4 & 87 \\
		
		& MoRPI-PINN
		& 4.2 & 3.5 & 1.0 & 0.7
		& 2.9 & 30 \\
		
		& PiDR (ours)
		& 4.0 & 3.4 & 0.3 & 0.3
		& 2.0 & 0.5 \\
		
		& PiDR-EKF (ours)
		& 4.0 & 3.4 & 0.3 & 0.3
		& 2.0 & -- \\

			\midrule
		
		\multirow{5}{*}{\textbf{FDE [m]}}
		& 2D INS
		& 6.7 & 9.4 & 11.4 & 13.0
		& 10.1 & 98 \\
		
		& MoRPI
		& 24.6 & 24.4 & 32.0 & 31.4
		& 28.1 & 99 \\
		
		& MoRPI-PINN
		& \textbf{0.1} & \textbf{0.1} & 1.8 & 1.8
		& 1.0 & 79 \\
		
		& PiDR (ours)
		& 0.3 & 0.1 & 0.2 & 0.3
		& 0.2 & 2 \\
		
		& PiDR-EKF (ours)
		& 0.2 & 0.1 & 0.2 & 0.3
		& 0.2 & -- \\
		\midrule
		\multirow{5}{*}{\textbf{TDE [\%]}}
		& 2D INS
		& 27 & 32 & 26 & 30
		& 28 & 37 \\
		
		& MoRPI
		& 65 & 65 & 74 & 73
		& 69 & 73 \\
		
		& MoRPI-PINN
		& 20 & 19 & 20 & 20
		& 20 & 8 \\
		
		& PiDR (ours)
		& 19 & 17 & 18 & 17
		& 18 & 0.1 \\
		
		& PiDR-EKF (ours)
		& 19 & 17 & 17 & 17
		& 18 & -- \\

		\bottomrule
	\end{tabularx}
\end{table*}

\noindent
The 2D INS suffers from substantial drift across all trajectories. 
This behavior is reflected in the high PRMSE of 8.1\,[m] and MATE of 7.6\,[m], along with a large TDE of 28\%, and an average FDE of 10.1\,[m]. These results confirm the well-known limitations of standalone INS in GNSS-denied environments.

\noindent 
However, as reported in Table~\ref{tab:ROSbot}, MoRPI consistently exhibits higher errors than the 2D INS baseline, with error magnitudes approximately twice those of 2D INS. Additionally, the performance of MoRPI degrades by at least 10\% on rectangular trajectories compared to circular trajectories, primarily due to abrupt changes in the yaw angle.

\noindent
The hybrid MoRPI-PINN approach significantly improves the trajectory prediction, reducing both the MATE and PRMSE to 2.9\,[m]. It further lowers the value of average TDE to 20\%, indicating improved trajectory shape preservation. Nevertheless, MoRPI-PINN exhibits outstanding endpoint accuracy, aligning perfectly with the accuracy of our approach.

\noindent
The PiDR framework incorporates the physical constraints into the training of neural network through the physics residual term $\mathscr{L}_{\text{phys}}(\vartheta)$. By enforcing the physics of ~\eqref{eq:pinn}, the model suppresses drifts and produces a prediction consistent with inertial motion. This improvement is evident across all reported metrics. 
PiDR demonstrates 70\% and 95\% improvements in PRMSE for circular and rectangular trajectories, respectively.
A similar trend is observed for the MATE, where PiDR reduces the average error to 2.0 [m], corresponding to an improvement of more than 73\% over conventional INS-based approaches.
Furthermore, PiDR reduces overall trajectory drift and achieves an average TDE of 18\%, indicating strong consistency over long-duration missions. 
However, FDE exhibits a different behavior: although PiDR attains an 89\% improvement over model-based baselines, its performance degrades for circular trajectories compared to MoRPI-PINN.
These findings demonstrate the advantages of incorporating physics during the learning process for robust inertial navigation.

\noindent For the mobile robot experiments, the incorporation of the EKF yields only marginal improvements over the PiDR results, with an average gain of approximately 1\% across the evaluated metrics. This limited improvement can be attributed to the fact that PiDR already produces highly accurate and smooth trajectory estimates, leaving little room for further refinement.

\noindent
From Table~\ref{tab:ROSbot}, it is evident that PiDR consistently achieves the lowest errors across all evaluated metrics and demonstrates strong robustness under planar motion for mobile robots.  

\subsubsection{AUV}
\begin{figure*}[!t]
\centering

\subfloat[Validation 1]{%
\includegraphics[width=0.95\textwidth]{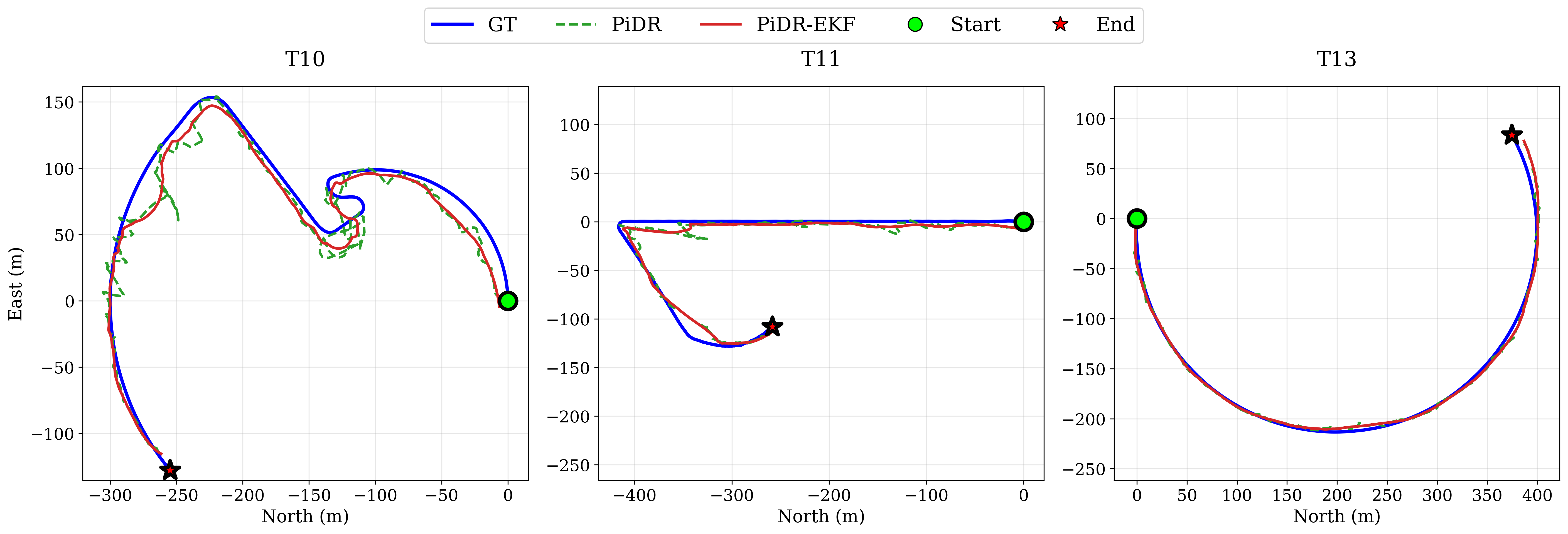}
\label{fig:traj_val1}
}

\vspace{8mm}

\subfloat[Validation 2]{%
\includegraphics[width=0.95\textwidth]{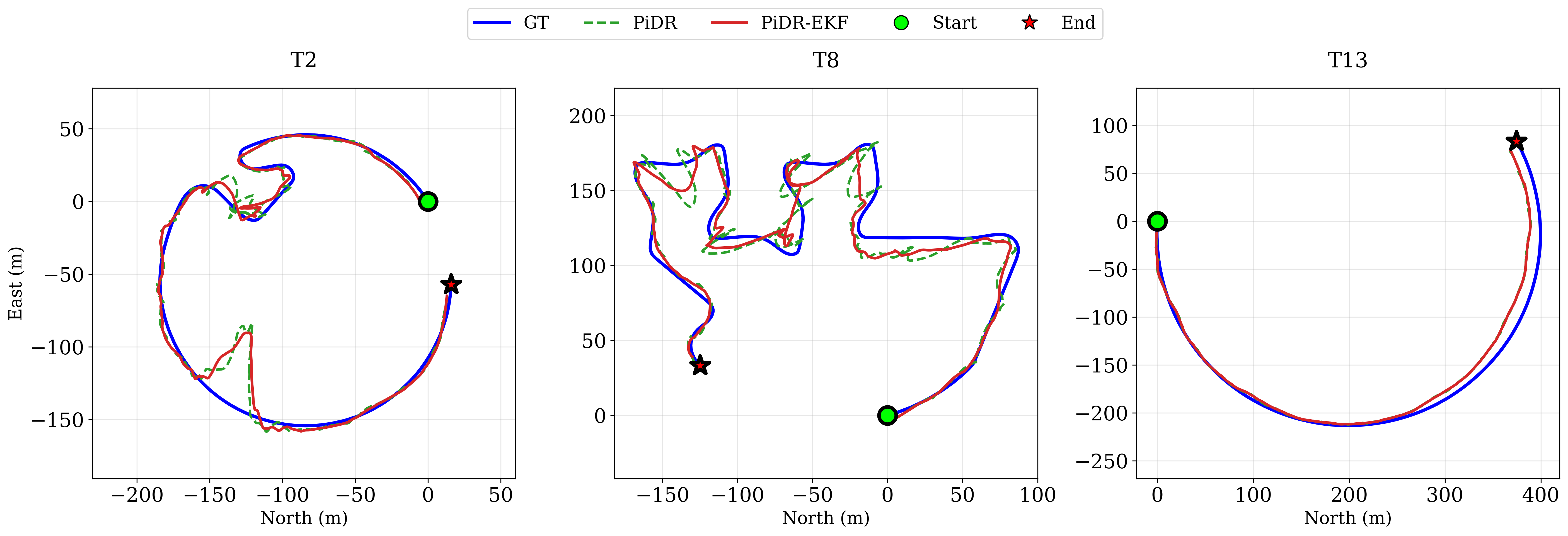}
\label{fig:traj_val2}
}

\vspace{8mm}

\subfloat[Validation 3]{%
\includegraphics[width=0.95\textwidth]{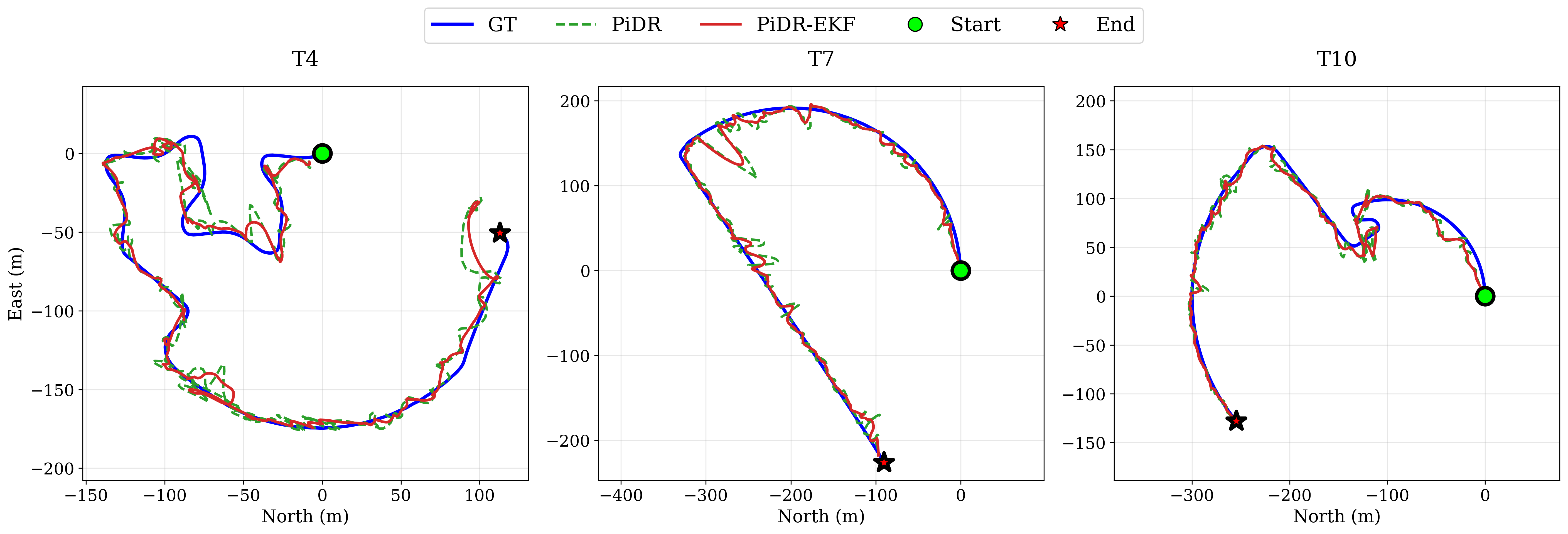}
\label{m}
}

\caption{Trajectory comparison for the three validation cases.}
\label{fig:traj_auv_test}
\end{figure*}

\noindent A top-view of the AUV's test trajectories are shown in Fig.~\ref{fig:traj_auv_test}. The 3D INS exhibits severe error accumulation, with an average PRMSE of 528.3~[m] and TDE exceeding 170\%, confirming the limitations of pure inertial navigation in underwater operations.

\begin{table*}[!ht]
	\centering
	\caption{Evaluation of the proposed PiDR and PiDR-EKF on the AUV trajectories. Results are averaged over AUV test trajectories Set-1.}
	\label{tab:metrics_comparison}
	\renewcommand{\arraystretch}{0.75}
	\begin{tabularx}{\textwidth}{l l
			*{3}{>{\centering\arraybackslash}X}
			>{\centering\arraybackslash}X
			>{\centering\arraybackslash}X}
		
		\toprule
		\textbf{Metric} & \textbf{Method}
		& \textbf{T10} & \textbf{T11} & \textbf{T13}
		& \textbf{Average} & \textbf{Improvement [\%] using PiDR-EKF} \\
		\midrule
		
		\multirow{5}{*}{\textbf{PRMSE [m]}}
		& 3D INS 
		& 545.3 & 529.9 & 511.6 & 528.3 & 97 \\
		
		& MoRPI 
		& 203.0 & 525.5 & 360.5 & 363.0 & 97 \\
		
		& MoRPI-PINN
		& 206.9 & 249.7 & 312.5 & 256.4 & 95 \\
		
		& PiDR (ours) 
		& 18.5 & 16.3 & 9.0 & 14.6 & 23 \\
		
		& PiDR-EKF (ours) 
		& 13.3 & 12.1 & 7.5 & 11.0 & -- \\
		
		\midrule
		\multirow{5}{*}{\textbf{MATE [m]}}
		& 3D INS 
		& 406.2 & 391.1 & 381.0 & 392.8 & 97 \\
		
		& MoRPI 
		& 175.2 & 515.3 & 254.9 & 315.1 & 97 \\
		
		& MoRPI-PINN
		& 196.3 & 229.2 & 277.0 & 234.2 & 96 \\
		
		& PiDR (ours) 
		& 12.1 & 10.7 & 7.1 & 10.0 & 15 \\
		
		& PiDR-EKF (ours) 
		& 10.5 & 9.1 & 6.2 & 8.6 & -- \\
		\midrule
		\multirow{5}{*}{\textbf{FDE [m]}}
		& 3D INS 
		& 1211.3 & 1248.8 & 1145.5 & 1201.9 & 99 \\
		
		& MoRPI 
		& 178.8 & 534.4 & 509.8 & 407.7 & 97 \\
		
		& MoRPI-PINN 
		& 229.9 & 226.4 & 435.6 & 297.3 & 95 \\
		
		& PiDR (ours) 
		& 15.8 & 6.9 & 17.2 & 13.3 & 8 \\
		
		& PiDR-EKF (ours) 
		& 15.9 & 6.7 & 14.2 & 12.3 & -- \\
		
		\midrule
		\multirow{5}{*}{\textbf{TDE (\%)}}
		& 3D INS 
		& 191.0 & 189.0 & 133.0 & 171.0 & 95 \\
		
		& MoRPI 
		& 44.0 & 98.0 & 133.0 & 91.7 & 86 \\
		
		& MoRPI-PINN 
		& 73.0 & 89.0 & 81.0 & 81.0 & 85 \\
		
		& PiDR (ours) 
		& 2.6 & 2.5 & 1.2 & 2.1 & 21 \\
		
		& PiDR-EKF (ours) 
		& 1.9 & 1.9 & 1.0 & 1.6 & -- \\

		\bottomrule
	\end{tabularx}
\end{table*}

\noindent
The MoRPI model achieves partial error reduction on certain trajectories (T10 and T13); however, its performance degrades substantially when the AUV starts diving into the deep water (T11), indicating limited generalization under varying motion dynamics. 
MoRPI-PINN improves accuracy by incorporating motion constraints, yet still accumulates significant drift over long and regular trajectories.

\noindent
In contrast, PiDR outperforms all baselines, achieving an improvement of 94\% for PRMSE. 
Additionally, the MATE is obtained in the range of 7--12\,[m], which is 3\% of the total trajectory length. Moreover, the TDE remains below 3\% for all trajectories, demonstrating the ability to maintain stable long-term underwater navigation with minimal drift.
Finally, the average FDE remains under 14\,[m] for all test trajectories. This is particularly significant for long-duration AUV missions, where terminal error directly impacts the goal of the mission.

\noindent For the AUV experiments, our PiDR-EKF model leads to a noticeable improvement. The proposed framework achieves average of 20\% improvement over the baseline PiDR results. 
To further validate the robustness, two additional cross-validation experiments were conducted using different train--test splits. These experiments resulted in improvements of approximately 7\% and 42\%, respectively. It demonstrates consistent performance gains across varying trajectory configurations and motion conditions. The observed improvements confirm that the PiDR--EKF framework is particularly beneficial in complex 3D underwater environments, where noise, bias, and nonlinear motion dynamics significantly affect inertial navigation accuracy.

\noindent All the obtained metrics are presented in Tables~\ref{tab:metrics_comparison} and \ref{tab:cross_validation}, with the numerical values and improvement percentage. These results demonstrate the effectiveness of the PiDR and PiDR-EKF models
to mitigate unbounded error growth inherent to pure inertial integration in underwater missions.

\begin{table*}[htbp]
\centering
\caption{Cross-validation results for PiDR and PiDR-EKF frameworks. 
}
\label{tab:cross_validation}
\begin{tabular}{c c c c  c c}
\hline
\textbf{Set No.} & \textbf{Train Set} & \textbf{Test Traj.} & \textbf{PIDR PRMSE} & \textbf{PiDR-EKF PRMSE}  & \textbf{Improvement (\%)} \\
\hline

\multirow{3}{*}{1} 
& \multirow{3}{*}{1,2,3,5,6,7,8} 
& 10 & 18.5 & 13.4  & 28 \\
& & 11 & 16.3 & 12.1  & 25 \\
& & 13 & 9.0  & 7.5   & 17 \\
\cline{3-6}
& & \textbf{Avg.} & -- & --  & \textbf{23} \\

\hline

\multirow{3}{*}{2} 
& \multirow{3}{*}{1,3,5,6,7,10,12} 
& 2  & 12.9 & 11.9 &  8 \\
& & 8  & 19.4 & 17.4 &  11 \\
& & 13 & 7.1  & 6.9   & 2 \\
\cline{3-6}
& & \textbf{Avg.} & -- & --  & \textbf{7} \\

\hline

\multirow{3}{*}{3} 
& \multirow{3}{*}{1,2,5,6,8,11,13} 
& 4  & 17.8 & 11.3 &  36 \\
& & 7  & 25.9 & 12.3 &  53 \\
& & 10 & 14.3 & 9.0  &  37 \\
\cline{3-6}
& & \textbf{Avg.} & -- & --  & \textbf{42} \\

\hline
\end{tabular}
\end{table*}

\subsection{Ablation Study}
\noindent To understand the contribution of each loss component, we analyze the role of the different loss terms in the training process.
Table~\ref{tab:ablation} presents an ablation and sensitivity analysis of the proposed PiDR framework on AUV test trajectories without EKF filtering. The results demonstrate that the proposed PiDR model, which combines both data and physics losses, achieves the lowest error. Removing key components, such as velocity and orientation supervision or reducing the number of collocation points, leads to noticeable performance degradation. Furthermore, the data-only and physics-only variants exhibit significantly higher errors. This underscores the requirement for dual enforcement of data fidelity and physical consistency. These findings confirm that the balanced incorporation of all the losses are critical for achieving robust inertial navigation performance.

\begin{table}[htbp]
\centering
\caption{Ablation and sensitivity analysis of the proposed PiDR framework. 
Results are averaged over AUV test trajectories Set-1 (T10, T11, and T13).}
\label{tab:ablation}
\footnotesize
\setlength{\tabcolsep}{4pt}
\begin{tabular}{p{3.8cm}ccc}
\toprule
\textbf{Model Variant} & \textbf{Data} & \textbf{Phys} & \textbf{MATE [m]} \\
\midrule
PiDR (ours) & $\checkmark$ & $\checkmark$ & \textbf{10.0} \\

Fixed-weight ($\lambda=1$) 
& $\checkmark$ & $\checkmark$ & 12.4 \\

Position loss-only 
& $\checkmark$ & $\checkmark$ & 14.5 \\

PiDR (700 collocation pts) 
& $\checkmark$ & $\checkmark$ & 17.4 \\

Data-only 
& $\checkmark$ & -- & 123.6 \\

Physics-only 
& -- & $\checkmark$ & 274.5 \\

Vel. \& orient. loss-only 
& $\checkmark$ & $\checkmark$ & 276.0 \\

\bottomrule
\end{tabular}
\end{table}

\subsection{Summary}
\noindent
This work demonstrates that the PiDR model provides a physics-consistent unified inertial navigation solution for both mobile robots and AUVs operating in GNSS-denied environments. Although the magnitude of error varies due to motion dynamics, varying mission profiles, and operational areas, the PiDR model yields relative improvements across both platforms.

\noindent
For ground robots, experimental results confirm that embedding inertial motion constraints directly within the learning architecture significantly reduces drift, stabilizes heading angle estimation, and enables accurate trajectory estimation using low-cost MEMS IMUs. However, the improvement of PiDR-EKF model is marginal in planar mobile robot scenarios. Finally, our model outperforms all baseline models and achieves an improvement of 11\%.

\noindent
In the underwater operations, PiDR achieves minimal accumulated drift, demonstrating its suitability for long-duration AUV missions in the absence of external aids. It is well known that IDR errors grow exponentially during critical maneuvers. As a result, trajectories with higher curvature yield larger ATE and PRMSE values, even when TDE remains stable. 
Although PiDR exhibits higher MATE values of up to 12 [m] in these challenging conditions, it still yields a 94\% improvement over the baseline models.
These results highlight the significance of physics-informed constraints in enabling stable and robust navigation for long-duration missions.
Furthermore, the EKF refinement effectively reduces noise, yielding more significant improvements in complex 3D AUV trajectories.

\noindent
Overall, Tables \ref{tab:ROSbot}, and \ref{tab:metrics_comparison}  demonstrate that PiDR outperforms all the baseline models across all evaluated metrics, validating the proposed PiDR framework as a reliable solution for inertial navigation tasks in GNSS-denied environments.

\noindent
In summary, INS models suffer from accumulated integration errors without external correction, making them unsuitable for extended trajectories with low-cost IMUs. While MoRPI constrains motion through geometric reconstruction, it lacks explicit enforcement of system dynamics, leading to increased errors during aggressive or critical maneuvering. Additionally, the drift grows rapidly during sudden changes in heading angle. Although MoRPI-PINN achieves very small endpoint errors for certain 2D trajectories, its performance degrades noticeably for 3D missions. Also, the existing models are platform-dependent.
In contrast, PiDR explicitly provides a unified inertial navigation solution for autonomous platforms operated in both 2D and 3D environments. Finally, integration of EKF with PiDR further improves the inertial navigation solutions in GNSS-denied environment.

\section{Conclusion} \label{sec:conclusion}
\noindent  In pure inertial navigation, low-cost inertial sensors cause rapid drift as their measurements contain noise and other error terms. To mitigate drift, we developed PiDR, a physics-informed framework for GNSS-denied environments and situations where external updates are unavailable. To further enhance the accuracy and reduce residual drift, an EKF-based refinement stage is incorporated to fuse PiDR predictions as external updates.
We demonstrated using a mobile robot, where the motion is planar and less affected by environmental disturbances, and using an AUV, where hydrodynamic disturbances and complex 3D dynamics are present, that our PiDR outperforms the baseline approach for pure inertial navigation.

\noindent Our PiDR model seamlessly integrates the DL approach and the dynamics of strapdown inertial navigation, using a composite loss function.
Specifically, PiDR employs a neural state representation coupled with physics-based residual enforcement at collocation points, allowing the model to learn navigation states in a continuous and physically meaningful manner across multiple platforms. It enables interpretability of the model and also retains the flexibility of DNN. 

\noindent PiDR is trained and tested on primary recorded data of 97 [min] and a path length of 2,983 [m] and cross-validated on additional AUV data of 132 [min].
To demonstrate its effectiveness, we compare it against GT and three other models. The results demonstrate that PiDR achieves more than 55\% improvements for the mobile robot and AUV. Moreover, PiDR-EKF again achieves a minimum improvement of 7\% over PiDR for AUVs.

\noindent PiDR is a robust, lightweight, yet effective algorithm for inertial navigation. However, our model has some potential limitations. Primarily, the incorporation of strict physical constraints may degrade the performance when the platform undergoes abrupt pattern changes due to some external factors. 
Secondarily, due to the composite loss function, the training of PiDR required efficient hardware, such as GPUs, to meet the computational demands. However, the real-time implementation is lightweight.
Therefore, this approach is suitable for platforms with limited resources.

\noindent
In summary, this research demonstrates that PiDR and PiDR-EKF are well-suited for navigation tasks in GNSS-denied environments.  It can be deployed on a wide range of autonomous platforms equipped with low-cost inertial sensors. It is viable for practical missions, such as underwater mapping, marine surveys,  search-and-rescue operations, security, and surveillance. Beyond performance gains, it also bridges the gap between purely data-driven models and model-based DR approaches.

\section*{Data Availability}
\noindent The dataset used in this study is publicly available at: 
\url{https://ansfl.marsci.haifa.ac.il/datasets}

\bibliographystyle{IEEEtran}
\bibliography{New_IEEEtran}

\end{document}